%% file: main.tex
\documentclass{article} % For LaTeX2e
\usepackage{iclr2021_conference,times}

\input{math_commands.tex}

\input{preamble}

\usepackage{xcolor,colortbl}

\usepackage{url}
\usepackage{hyperref}  
\hypersetup{colorlinks=true,citecolor=blue,linkcolor=blue}

\usepackage{comment}
\usepackage{amsmath}
\usepackage{amsfonts}
\usepackage{bm}
\usepackage{xfrac}
\newcommand{\norm}[1]{\left\lVert#1\right\rVert}

\usepackage{graphicx}
\usepackage{float}
\usepackage{booktabs}
\usepackage{tabularx}
\graphicspath{ {./images/} }

\usepackage[skip=4pt]{caption}

\usepackage{algorithm}
\usepackage{algpseudocode}

\newcommand{\Cam}[1]{}

\title{Exact Representation of Sparse Networks with Symmetric Nonnegative Embeddings}
\usepackage{authblk}
\author[1]{Sudhanshu Chanpuriya}
\author[2]{Ryan A. Rossi}
\author[2]{Anup Rao}
\author[2]{Tung Mai}
\author[2]{\\Nedim Lipka}
\author[2]{Zhao Song}
\author[1]{Cameron Musco}
\affil[1]{University of Massachusetts Amherst, \texttt{\{schanpuriya,cmusco\}@cs.umass.edu}} 
\affil[2]{Adobe Research, \texttt{\{ryrossi,anuprao,tumai,lipka,zsong\}@adobe.com}}

\usepackage[preprint]{neurips_2022}

\begin{document}

\maketitle

\begin{abstract}
Many models for undirected graphs are based on factorizing the graph's adjacency matrix; these models find a vector representation of each node such that the predicted probability of a link between two nodes increases with the similarity (dot product) of their associated vectors. Recent work has shown that these models are unable to capture key structures in real-world graphs, particularly heterophilous structures, wherein links occur between dissimilar nodes.
In contrast, a factorization with two vectors per node, based on logistic principal components analysis (LPCA), has been proven not only to represent such structures, but also to provide exact low-rank factorization of any graph with bounded max degree.
However, this bound has limited applicability to real-world networks, which often have power law degree distributions with high max degree. Further, the LPCA model lacks interpretability since its asymmetric factorization does not reflect the undirectedness of the graph.

We address the above issues in two ways. First, we prove a new bound for the LPCA model in terms of \emph{arboricity} rather than max degree; this greatly increases the bound's applicability to many sparse real-world networks.
Second, we propose an alternative graph model whose factorization is symmetric and nonnegative, which allows for link predictions to be interpreted in terms of node clusters. We show that the bounds for exact representation in the LPCA model extend to our new model.
On the empirical side, our model is optimized effectively on real-world graphs with gradient descent on a cross-entropy loss. We demonstrate its effectiveness on a variety of foundational tasks, such as community detection and link prediction.
\end{abstract}

% \makeatletter
% \renewcommand\paragraph{\@startsection{paragraph}{4}{\z@}{3.25ex \@plus1ex \@minus.2ex}{-1em}{\normalfont\normalsize\bfseries}}
% \makeatother

\section{Introduction}\label{sec:intro}

Graphs naturally arise in data from a variety of fields including sociology~\citep{mason2007graph}, biology~\citep{scott1988social}, and computer networking~\citep{bonato2004survey}.
A key underlying task in machine learning for graph data is forming models of graphs which can predict edges between nodes, form useful representations of nodes, and reveal interpretable structure in the graph, such as detecting clusters of nodes.
Many graph models fall under the framework of edge-independent graph generative models, which can output the probabilities of edges existing between any pair of nodes. The parameters of such models can be trained iteratively on the network, or some fraction of the network which is known, in the link prediction task, i.e., by minimizing a predictive loss.
To choose among these models, one must consider two criteria: 1) whether the model can express structures of interest in the graph, 2) whether the model expresses these structure in an interpretable way.

% \Cam{I think we don't shouldn't introduce the concept of `edge-independence' or focus too much on 'generative models' above and just can say that many models predict edge probabilities. I also think we should highlight the 3 considerations (maybe as a numbered list like (1) blah, (2) blah, (3) blah, as expressivity, intepretability, and trainablity or optimizability or whatever you want to call it. Since our model is meant to do well in all 3.}

\vspace{-2.5pt}
\paragraph{Expressiveness of low-dimensional embeddings} 
As real-world graphs are high-dimensional objects, graph models generally compress information about the graph.
Such models are exemplified by the family of dot product models, which associate each node with a real-valued ``embedding'' vector; the predicted probability of a link between two nodes increases with the similarity of their embedding vectors.
These models can alternatively be seen as factorizing the graph's adjacency matrix to approximate it with a low-rank matrix.
Recent work of~\cite{seshadhri2020impossibility} has shown that dot product models are limited in their ability to model  common structures in real-world graphs, such as triangles incident only on low-degree nodes. In response, \citet{chanpuriya2020node} showed that with the logistic principal components analysis (LPCA) model, which has two embeddings per node (i.e., using the dot product of the `left' embedding of one node and the `right' embedding of another), not only can such structures be represented, but further, any graph can be exactly represented with embedding vectors whose lengths are linear in the max degree of the graph.
There are two keys to this result. First is the presence of a nonlinear linking function in the LPCA model; since adjacency matrices are generally not low-rank, exact low-rank factorization is generally impossible without a linking function. Second is that having two embeddings rather than one allows for expression of non-positive semidefinite (PSD) matrices. As discussed in 
\citet{peysakhovich2021attract} that the single-embedding models can only represent PSD matrices precludes representation of `heterophilous' structures in graphs; heterophilous structures are those wherein dissimilar nodes are linked, in contrast to more intuitive `homophilous' linking between similar nodes.
% Gives a more general understanding
% \Cam{How does the \cite{peysakhovich2021attract} paper exactly relate to ours? Like are they saying that triangle dense, sparse graphs must have heterophilous structure, and this is why they can't be represented by dot product models? In general, we are claiming that dot product models are bad because they can't represent heterophilous structure. But they also can't represent high triangle density. Are these two different things or one in the same? Does the new model over come both?}

\vspace{-2.5pt}
\paragraph{Interpretability and node clustering} Beyond being able to capture a given network accurately, it is often desirable for a graph model to form interpretable representations of nodes and to produce edge probabilities in an interpretable fashion. Dot product models can achieve this by restricting the node embeddings to be nonnegative. Nonnegative factorization has long been used to decompose data into parts~\citep{donoho2004does}. In the context of graphs, this entails decomposing the set of nodes of the network into clusters or communities. In particular, each entry of the nonnegative embedding vector of a node represents the intensity with which the node participates in a community.
This allows the edge probabilities output by dot product models to be interpretable in terms of coparticipation in communities.
% \Cam{Is tehre prior work we can cite taht talks about why non-negative embeddings are more interpretable for graphs? E.g. is this a selling point of BigClam?}
Depending on the model, these vectors may have restrictions such as a sum-to-one requirement, meaning the node is assigned a categorical distribution over communities.
The least restrictive and most expressive case is that of soft assignments to overlapping communities, where the entries can vary totally independently.
In such models, which include the \textsc{BigClam} model of \cite{yang2013overlapping}, the output of the dot product may be mapped through a nonlinear link function (as in LPCA) to produce a probability for each edge, i.e., to ensure the values lie in $[0,1]$.
% This link function ideally also facilitates straightforward interpretation.
% \Cam{Maybe not worth addressing but it is a bit confusing what non-negativity add. If you have negative entries, maybe it just means that that node is adverse to people in that community. Still is pretty interpretable.}

\begin{figure*}[h!]
\centering
\includegraphics[width=0.5\textwidth]{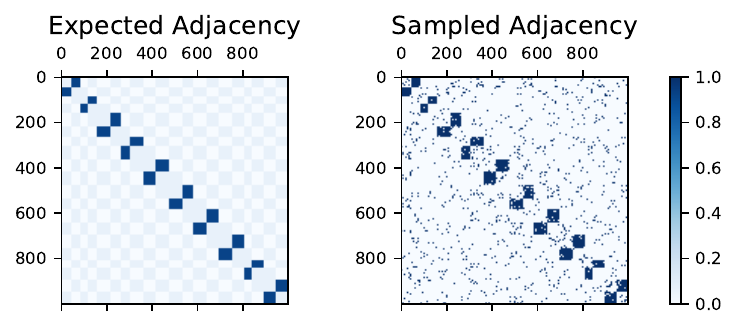}
\vspace{-3.5pt}
\caption{The motivating synthetic graph. The expected adjacency matrix (left) and the sampled matrix (right); the latter is passed to the training algorithms. The network is approximately a union of ten bipartite graphs, each of which correspond to men and women in one of the ten cities.}
\label{fig:synth_true}
\end{figure*}

\begin{figure*}[h!]
\centering
\includegraphics[width=0.7\textwidth]{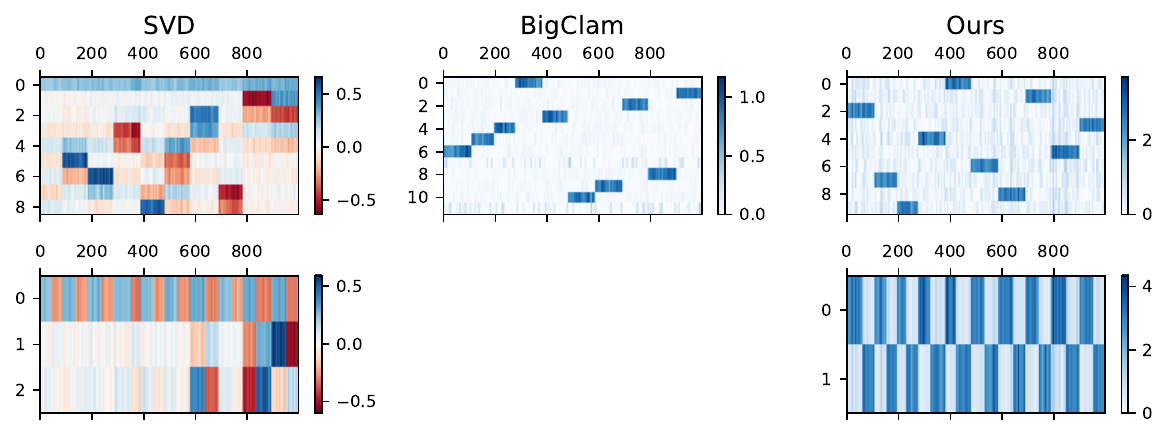}
\vspace{-3.5pt}
\caption{Factors resulting from decomposition of the motivating synthetic graph of Figure~\ref{fig:synth_true} with the three models, using 12 communities or singular vectors. The top/bottom rows represent the positive/negative eigenvalues corresponding to homophilous/heterophilous communities (note that \textsc{BigClam} does not include the latter). The homophilous factors from \textsc{BigClam} and our model reflect the 10 cities, and the heterophilous factor from our model reflect men and women. The factors from SVD are harder to interpret. Note that the order of the communities in the factors is arbitrary.}
\label{fig:synth_feats}
\end{figure*}

\vspace{-2.5pt}
\paragraph{Heterophily: Motivating example} To demonstrate how heterophily can manifest in networks, as well as how models which assume homophily can fail to represent such networks, we provide a simple synthetic example. Suppose we have a graph of matches between users of a mostly heterosexual dating app, and the users each come from one of ten cities. Members from the same city are likely to match with each other; this typifies homophily, wherein links occur between similar nodes. Furthermore, users having the same gender are are unlikely to match with each other; this typifies heterophily. Figure~\ref{fig:synth_true} shows an instantiation of such an adjacency matrix with $1000$ nodes, which are randomly assigned to man or woman and to one of the ten cities. 
We recreate this network with our proposed embedding model and with \textsc{BigClam}, which explicitly assumes homophily. 
We also compare with the SVD of the adjacency matrix, which outputs the best (lowest Frobenius error) low-rank approximation that is possible without a nonlinear linking function. 
% Since SVD lacks nonnegativity constraints on the factors, we do not expect intepretability. 
In Figure~\ref{fig:synth_true}, we show how \textsc{BigClam} captures only the ten communities based on city, i.e., only the homophilous structure, and fails to capture the heterophilous distinction between men and women. We also plot the error of the reconstructions as the embedding length increases.
% Since there are $10\cdot 2 = 20$ different kinds of nodes, the expected adjacency matrix is rank-$20$.
There are $10\cdot 2 = 20$ different kinds of nodes, meaning the expected adjacency matrix is rank-$20$, and our model maintains the lowest error up to this embedding length; by contrast, \textsc{BigClam} is unable to decrease error after capturing city information with length-$10$ embeddings.
In Figure~\ref{fig:synth_feats}, we visualize the features generated by the three methods, i.e., the factors returned by each factorization. Our model's factors captures the relevant latent structure in an interpretable way. By contrast, SVD's factors are harder to interpret, and \textsc{BigClam} does not represent the heterophilous structure.

% We propose the first factorization-based graph model that a) is expressive enough to capture heterophily, b) produces nonnegative node representations, which allow link predictions to be interpreted in terms of node clusters, and c) outputs edge probabilities and optimizes effectively on real-world graphs with gradient descent on a cross-entropy loss. 

\vspace{-2.5pt}
\paragraph{Summary of main contributions}
The key contributions of this work are as follows:
\begin{itemize}[itemsep=2pt,topsep=2pt]
    \item We prove that the LPCA model admits exact low-rank factorizations of graphs with bounded \emph{arboricity}, which is the minimum number of forests into which a graph's edges can be partitioned. By the Nash-Williams theorem, arboricity is a measure of a graph's density in that, letting $S$ denote an induced subgraph and $n_S$ and $m_S$ denote the number of nodes and edges in $S$, arboricity is the maximum over all subgraphs $S$ of $\ceil{ \tfrac{m_S}{n_S-1} }$. Our result is more applicable to real-world graphs than the prior one for graphs with bounded max degree, since sparsity is a common feature of real networks, whereas low max degree is not.
    \item We introduce a graph model which is both highly expressive and interpretable. Our model incorporates two embeddings per node and a nonlinear linking function, and hence is able to express both heterophily and overlapping communities. At the same time, our model is based on symmetric nonnegative matrix factorization, so it outputs link probabilities which are interpretable in terms of the communities it detects.
    \item We show how any graph with a low-rank factorization in the LPCA model also admits a low-rank factorization in our community-based model. This means that the guarantees on low-rank representation for bounded max degree and arboricity also apply to our model.
    % %
    % \item {\color{red}We provide a scheme for initialization of the nonnegative factors using the arbitrary real factors generated by logistic PCA. We show theoretically how a graph which is represented exactly by LPCA can also be represented exactly by our model.}
    % %
    % \item We prove that, with a small number of communities, our model can exactly represent a natural class of graphs which exhibits both heterophily and overlapping communities.
    %
    \item In experiments, we show that our method is competitive with and often outperforms other comparable models on real-world graphs in terms of representing the network, doing interpretable link prediction, and detecting communities that align with ground-truth. 
    % \Cam{Are we not able to say it outperforms? Or at least outperforms in some cases?}
\end{itemize}

% We propose the first factorization-based graph model that a) is expressive enough to capture heterophily, b) produces nonnegative node representations, which allow link predictions to be interpreted in terms of node clusters, and c) outputs edge probabilities and optimizes effectively on real-world graphs with gradient descent on a cross-entropy loss.

% \paragraph{Main contributions} We provide a scheme for initialization of the nonnegative factors using the arbitrary real factors generated by logistic PCA. We show theoretically how a graph which is represented exactly by LPCA can also be represented exactly by our model.We show theoretically that, with a small number of communities, our model can exactly represent a natural class of graphs which exhibits both heterophily and overlapping communities. With experiments, we show that our algorithm is competitive on real-world graphs in terms of representing the network, doing interpretable link prediction, and detecting communities that align with ground-truth.

\vspace{-3.5pt}
\begin{figure*}[h!]
\centering
\includegraphics[width=0.7\textwidth]{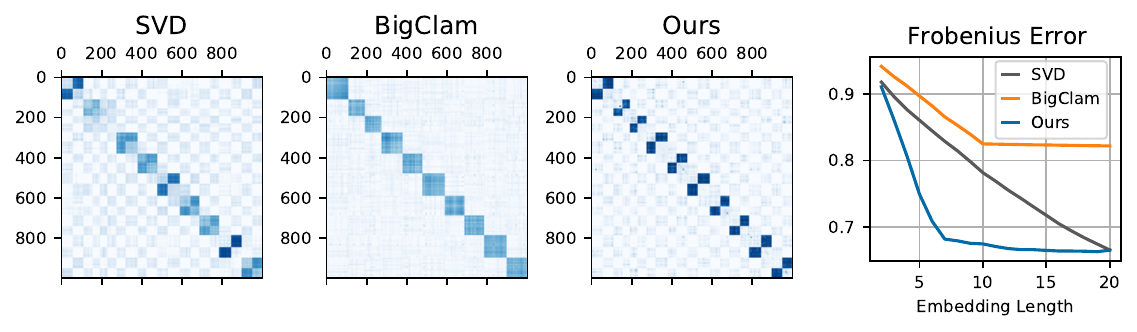}
\vspace{-3.5pt}
\caption{Left: Reconstructions of the motivating synthetic graph of Figure~\ref{fig:synth_true} with SVD, \textsc{BigClam}, and our model, using 12 communities or singular vectors. Note the lack of the small diagonal structure in \textsc{BigClam}'s reconstruction; this corresponds to its inability to capture the heterophilous interaction between men and women. Right: Frobenius error when reconstructing the motivating synthetic graph of Figure~\ref{fig:synth_true} with SVD, \textsc{BigClam}, and our model, as the embedding length is varied. The error is normalized by the sum of the true adjacency matrix (i.e., twice the number of edges).}
\label{fig:synth_recons}
\end{figure*}

\vspace{-3.5pt}

\section{Community-Based Graph Factorization Model}\label{sec:model}

Consider the set of undirected, unweighted graphs on $n$ nodes, i.e., the set of graphs with symmetric adjacency matrices in $\{0,1\}^{n \times n}$.
We propose an edge-independent generative model for such graphs. Given nonnegative parameter matrices $\mB \in \mathbb{R}_+^{n \times k_B}$ and $\mC \in \mathbb{R}_+^{n \times k_C}$, we set the probability of an edge existing between nodes $i$ and $j$ to be the $(i,j)$-th entry of matrix $\tilde{\mA}$:
\begin{align} \label{eqn:bbcc}
\tilde{\mA} := \sigma( \mB \mB^\top - \mC \mC^\top ) ,
\end{align}
where $\sigma$ is the logistic function. Here $k_B$, $k_C$ are the number of homophilous/heterophilous clusters. Intuitively, if $\vb_i \in \mathbb{R}_+^{k_B}$ is the $i$-th row of matrix $\mB$, then $\vb_i$ is the affinity of node $i$ to each of the $k_B$ homophilous communities. Similarly, $\vc_i \in \mathbb{R}_+^{k_C}$ is the affinity of node $i$ to the $k_C$ heterophilous communities. As an equivalent statement, for each pair of nodes $i$ and $j$, $\tilde{\mA}_{i,j} := \sigma( \vb_i \vb_j^\top - \vc_i \vc_j^\top )$. 
We will soon discuss the precise interpretation of this model, but the idea is roughly similar to the attract-repel framework of \citet{peysakhovich2021attract}. When nodes $i$ and $j$ have similar `attractive' $\vb$ embeddings, i.e., when $\vb_i \vb_j^\top$ is high, the likelihood of an edge between them increases, hence why the $\mB$ factor is homophilous. By contrast, the $\mC$ factor is `repulsive'/heterophilous since, when $\vc_i \vc_j^\top$ is high, the likelihood of an edge between $i$ and $j$ decreases.

% The edge probabilities output by this model have an intuitive interpretation. Recall that there are bijections between probability $p \in [0,1]$, odds $o=\tfrac{p}{1-p} \in [0,\infty)$, and logit $\ell = \log(o)\in (-\infty,+\infty)$. The logit of the link probability between nodes $i$ and $j$ in this model is $\vb_i \vb_j^\top - \vc_i \vc_j^\top$. The contribution of the $k^\text{th}$
% , which is a summation of terms $\vv_{ic} \vv_{jc} \mW_{cc}$ over all communities $c \in [k]$. If the nodes both fully participate in community $c$, that is, $\vv_{ic} = \vv_{jc} = 1$, then the edge logit is changed by $\mW_{cc}$ starting from a baseline of $0$, or equivalently, the odds of an edge is multiplied by $\exp(\mW_{cc})$ starting from a baseline odds of $1$; if either of the nodes participates only partially in community $c$, then the change in logit and odds is accordingly prorated. Homophily and heterophily also have a clear interpretation in this model: homophilous communities, which are expressed in $\mB$, are those with $\mW_{cc} > 0$, where two nodes both participating in the community increases the odds of a link, whereas communities with $\mW_{cc} < 0$, which are expressed in $\mC$, are heterophilous, and coparticipation decreases the odds of a link.

% \Cam{Its not really clear what homophilous and heterophilous communities are here. Should we mention up front that this is similar to the atract-repel framework of Bottou?}

\paragraph{Alternate expression} We note that the model above can also be expressed in a form which normalizes cluster assignments and is more compact, in that it combines the homophilous and heterophilous cluster assignments. Instead of $\mB$ and $\mC$, this form uses a matrix $\mV \in [0,1]^{n \times k}$ and a diagonal matrix $\mW\in\mathbb{R}^{k \times k}$, where $k=k_B+k_C$ is the total number of clusters.
In particular, let $\vm_B$ and $\vm_C$ be the vectors containing the maximums of each column of $\mB$ and $\mC$. By setting
\begin{align}
\label{eqn:bbcc_to_vwv}
\begin{split}
    \mV &= \begin{pmatrix} \mB \times \diag\left(\vm_B^{-1}\right); & \mC \times \diag\left(\vm_C^{-1}\right) \end{pmatrix} \\
    \mW &= \diag\left( \begin{pmatrix} 
        +\vm_B^{2}; & -\vm_C^{2}
    \end{pmatrix} \right) ,
\end{split}
\end{align}
the constraint on $\mV$ is satisfied. Further, $\mV \mW \mV^\top = \mB \mB^\top - \mC \mC^\top$, so
\begin{align} \label{eqn:vwv}
\tilde{\mA} := \sigma( \mB \mB^\top - \mC \mC^\top ) = \sigma( \mV \mW \mV^\top ).
\end{align}
Here, if $\vv_i \in [0,1]^k$ is the $i$-th row of matrix $\mV$, then $\vv_i$ is the soft (normalized) assignment of node $i$ to the $k$ communities. The diagonal entries of $\mW$ represent the strength of the homophily (if positive) or heterophily (if negative) of the communities.
For each entry, $\tilde{\mA}_{i,j} = \sigma( \vv_i \mW \vv_j^\top )$. We use these two forms interchangeably throughout this work.

\paragraph{Interpretation}
The edge probabilities output by this model have an intuitive interpretation. Recall that there are bijections between probability $p \in [0,1]$, odds $o=\tfrac{p}{1-p} \in [0,\infty)$, and logit $\ell = \log(o)\in (-\infty,+\infty)$. The logit of the link probability between nodes $i$ and $j$ is $\vv_i^\top \mW \vv_j$, which is a summation of terms $\vv_{ic} \vv_{jc} \mW_{cc}$ over all communities $c \in [k]$. If the nodes both fully participate in community $c$, that is, $\vv_{ic} = \vv_{jc} = 1$, then the edge logit is changed by $\mW_{cc}$ starting from a baseline of $0$, or equivalently, the odds of an edge is multiplied by $\exp(\mW_{cc})$ starting from a baseline odds of $1$; if either of the nodes participates only partially in community $c$, then the change in logit and odds is accordingly prorated. Homophily and heterophily also have a clear interpretation in this model: homophilous communities, which are expressed in $\mB$, are those with $\mW_{cc} > 0$, where two nodes both participating in the community increases the odds of a link, whereas communities with $\mW_{cc} < 0$, which are expressed in $\mC$, are heterophilous, and coparticipation decreases the odds of a link.

\section{Related Work} \label{sec:related-work}

\paragraph{Community detection via interpretable factorizations} There is extensive prior work on the community detection / node clustering problem~\citep{schaeffer2007graph,aggarwal2010survey,nascimento2011spectral}, perhaps the most well-known being the normalized cuts algorithm of \citet{shi2000normalized}, which produces a clustering based on the entrywise signs of an eigenvector of the graph Laplacian matrix. However, the clustering algorithms which are most relevant to our work are those based on non-negative matrix factorization (NMF)~\citep{lee1999learning,berry2007algorithms,wang2012nonnegative,gillis2020nonnegative}. 
One such algorithm is that of \citet{yu2005soft}, which approximately factors a graph's adjacency matrix $\mA \in \{0,1\}^{n \times n}$ into two positive matrices $\mH$ and $\bm{\Lambda}$, where $\mH \in \mathbb{R}_+^{n \times k}$ is left-stochastic (i.e. each of its columns sums to $1$) and $\bm{\Lambda} \in \mathbb{R}_+^{k \times k}$ is diagonal, such that $\mH \bm{\Lambda} \mH^{\top} \approx \mA$. 
Here $\mH$ represents a soft clustering of the $n$ nodes into $k$ clusters, while the diagonal entries of $\bm{\Lambda}$ represent the prevalence of edges within clusters. 
Note the similarity of the factorization to our model, save for the lack of a nonlinearity.
Other NMF approaches include those of \citet{ding2008nonnegative}, \citet{yang2012clustering}, \citet{kuang2012symmetric}, and \citet{kuang2015symnmf} (\textsc{SymNMF}). 
% \Cam{I think some of this should be moved up to the intro to motivate what we are doing. See my comment there.}

\paragraph{Modeling heterophily} Much of the existing work on graph models has an underlying assumption of network homophily~\citep{johnson2010entropic, noldus2015assortativity}.
% ; homophily, sometimes called assortativity, is obeyed when an edge is more likely to appear between two nodes if they represent similar entities.
% In the context of node embedding, this assumption often means that clusters are assigned so that edges tend to appear between nodes in the same cluster. These approaches may not generalize for graphs with heterophily and could return poor clusterings in such cases. For example, clustering nodes of a nearly bipartite graph into two groups under a homophilous model may not return the natural clustering. 
There has been significant recent interest in the limitations of graph neural network (GNN) models~\citep{duvenaud2015convolutional,kipf2016semi,hamilton2017inductive} at addressing network heterophily~\citep{nt2019revisiting,zhu2020beyond}, as well as proposed solutions \citep{pei2020geom,yan2021two}, but relatively less work for more fundamental models such as those for clustering.
Some existing NMF approaches to clustering do naturally model heterophilous structure in networks.
% The model of \citet{nourbakhsh2014matrix}, for example, is similar to that of \cite{yu2005soft}, but allows the cluster affinity matrix $\bm{\Lambda}$ to be non-diagonal; this allows for inter-cluster edge affinity to exceed intra-cluster edge affinity, so heterophily can arise in this model, though it is not a focus of their work. 
For example, the model of \citet{miller2009nonparametric} is similar to ours and also allows for heterophily, though it restricts the cluster assignment matrix $\mV$ to be binary; additionally, their training algorithm is not based on gradient descent as ours is, and it does not scale to larger networks.
More recently, \citet{peysakhovich2021attract} propose a decomposition of the form $\mA \approx \mD + \mB \mB^\top - \mC \mC^\top$, where $\mD \in \mathbb{R}^{n \times n}$ is diagonal and $\mB,\mC \in \mathbb{R}^{n \times k}$ are low-rank.
% ; the authors discuss how, interestingly, this model separates the homophilous and heterophilous structure into different factors, namely $\mB$ and $\mC$.
Note that their decomposition does not include a nonlinear linking function, and their work does not pursue a clustering interpretation or investigate setting the factors $\mB$ and $\mC$ to be nonnegative.
% , and hence the entries of its reconstructed adjacency matrix are not restricted to be probabilities.
% , which precludes use of a cross-entropy gradient descent algorithm like ours for fitting the model.

\paragraph{Overlapping communities and exact embeddings} Many models discussed above focus on the single-label clustering task and thus involve highly-constrained factorizations (e.g., sum-to-one conditions). We are interested in the closely related but distinct task of multi-label clustering, also known as overlapping community detection~\citep{xie2013overlapping, javed2018community}, which involves less constrained, more expressive factorizations.
The \textsc{BigClam} algorithm of \citet{yang2013overlapping} uses the following generative model for this task: the probability of a link between two nodes $i$ and $j$ is given by $1 - \exp(-\bm{f}_i \cdot \bm{f}_j)$, where $\bm{f}_i, \bm{f}_j \in \mathbb{R}_+^k$ represent the intensities with which the nodes participate in each of the $k$ communities.
Note that \textsc{BigClam} assumes strict homophily of the communities: two nodes participating in the same community always increases the probability of a link.
However, this model allows for expression of very dense intersections of communities, which the authors observe is generally a characteristic of real-world networks.
To ensure that output entries are probabilities, \textsc{BigClam}'s factorization includes a nonlinear linking function (namely, $f(x)=1-e^x$), like our model and LPCA. Recent work outside clustering and community detection on graph generative models~\citep{rendsburg2020netgan,chanpuriya2020node} suggests that incorporating a linking function can greatly increase the expressiveness of factorization-based graph models, to the point of being able to exactly represent a graph.
This adds to a growing body of literature on expressiveness guarantees for embeddings on relational data~\citep{sala2018representation,bhattacharjee2020relations,boratko2021capacity}.
Most relevant to our work, as previously discussed, \citet{chanpuriya2020node} provide a guarantee for exact low-rank representation of graphs with bounded max degree when using the LPCA factorization model. In this work, we provide a new such guarantee, except for bounded arboricity, which is more applicable to real-world networks, and extend these guarantees to our community-based factorization.

% \Cam{I think should add some citation to work on boxicity and sphericity which are basically exacty embedding questions. Dongxu's paper migth be a good place to look for links: \url{https://proceedings.neurips.cc/paper/2021/file/88d25099b103efd638163ecb40a55589-Paper.pdf}}

\section{Theoretical Results} \label{sec:theory}

We first restate the main result from \cite{chanpuriya2020node} on exact representation of graphs with bounded max degree using the logistic principal components analysis (LPCA) model, which reconstructs a graph $\mA \in \{0,1\}^{n \times n}$ using logit factors $\mX,\mY \in \mathbb{R}^{n \times k}$ via
\begin{equation}~\label{eqn:lpca_model}
    \mA \approx \sigma(\mX \mY^\top).
\end{equation}
Note that unlike our community-based factorization, the factors of the LPCA model are not nonnegative, and the factorization does not reflect the symmetry of the undirected graph's adjacency matrix. Regardless of the model's interpretability, the following theorem provides a significant guarantee on its expressiveness. We use the following notation: given a matrix $\mM$, let $H(\mM)$ denote the matrix resulting from entrywise application of the Heaviside step function to $\mM$, that is, setting all positive entries to $1$, negative entries to $0$, and zero entries to $\sfrac{1}{2}$.
\begin{theorem}[Exact LPCA Factorization for Bounded-Degree Graphs, \cite{chanpuriya2020node}]\label{thm:lpca_exact_maxdeg} Let $\mA \in \{0,1\}^{n \times n}$ be the adjacency matrix of a graph $G$ with maximum degree $c$. Then there exist matrices $\mX,\mY \in \R^{n \times (2c+1)}$ such that $\mA = H(\mX \mY^\top)$.
\end{theorem}
%
% In other words, the entrywise signs of the product $\mX\mY^\top$
This corresponds to arbitrarily small approximation error in the LPCA model (Equation~\ref{eqn:lpca_model}) because, provided such factors $\mX,\mY$ for some graph $\mA$, we have that $\lim_{s \to \infty} \sigma\left( s \mX \mY^\top \right) = H(\mX \mY^\top ) = \mA$.
% \begin{equation*}
%     \lim_{s \to \infty} \sigma\left( s \mX \mY^\top \right) = H(\mX \mY^\top ) = \mA,
% \end{equation*}
That is, we can scale the factors larger to reduce the error to an arbitrary extent.

We expand on this result in two ways.
First, give  a new bound for exact embedding in terms of arboricity, rather than max degree. This significantly increases the applicability to real-world networks, which often are sparse (i.e., low arboricity) and have right-skewed degree distributions (i.e., high max degree).
% In Appendix~\ref{app:datasets}, we give five examples of real-world datasets, their max degrees, and upper bounds on their arboricities.
Second, we show that any rank-$k$ LPCA factorization can be converted to our model's symmetric nonnegative factorization with $O(k)$ communities. This extends the guarantees on the LPCA model's power for exact representation of graphs, both the prior guarantee in terms of max degree and our new one in terms of arboricity, to our community-based model as well.
In Appendix~\ref{app:cot}, we also introduce an example of a natural family of graphs - Community Overlap Threshold (COT) graphs - for which our model's community-based factorization not only exactly represents the graph, but also must capture some latent structure to do so with sufficiently low embedding dimensionality.

\paragraph{Arboricity bound for exact representation} We will use the following well-known fact: the rank of the entrywise product of two matrices is at most the product of their individual ranks, that is,% $\text{rank}(\mX \circ \mY) \leq \text{rank}(\mX) \cdot \text{rank}(\mY)$.
\begin{equation*}
    \text{rank}(\mX \circ \mY) \leq \text{rank}(\mX) \cdot \text{rank}(\mY).
\end{equation*}

\begin{theorem}[Exact LPCA Factorization for Bounded-Arboricity Graphs]\label{thm:lpca_exact_maxdeg_arboricity} Let $\mA \in \{0,1\}^{n \times n}$ be the adjacency matrix of an undirected graph $G$ with arboricity $\alpha$. Then there exist embeddings $\mX,\mY \in \R^{n \times (4\alpha^2+1)}$ such that $\mA = H(\mX \mY^\top)$.
\end{theorem}
\begin{proof}
Let the undirected graph $\mA$ have arboricity $\alpha$, i.e., the edges can be partitioned into $\alpha$ forests. We produce a directed graph $\mB$ from $\mA$ by orienting the edges in these forests so that each node's edges point towards its children. Now $\mA = \mB + \mB^\top$, and every node in $\mB$ has in-degree at most $\alpha$.

Let $\mV \in \R^{n \times 2\alpha}$ be the Vandermonde matrix with $\mV_{t,j} = t^{j-1}$. For any $\vc \in \R^{2\alpha}$, $[\mV \vc](t) = \sum_{j = 1}^{2\alpha} \vc({j}) \cdot t^{j-1}$, that is, $\mV \vc \in \R^{n}$ is a degree-$(2\alpha)$ polynomial with coefficients $\vc$ evaluated at the integers $t \in [n] = \{1,\dots,n\}$.
Let $\vb_i$ be the $i^{\text{th}}$ column of $\mB$. 
We seek to construct a polynomial such that for $t$ with $\vb_i(t) = 1$, $[\mV \vc_i](t) = 0$, and $[\mV \vc_i](t) < 0$ elsewhere; that is, when inputting an index $t \in [n]$ such that the $t^{\text{th}}$ node is an in-neighbor of the $i^{\text{th}}$ node, we want the polynomial to output $0$, and for all other indices in $[n]$, we want it to have a negative output. 
Letting $N(i)$ denote the in-neighbors of the $i^\text{th}$ node, a simple instantiation of such a polynomial in $t$ is $-1 \cdot \prod_{j \in N(i)} (t - j)^2$. Note that since all nodes have in-degree at most $\alpha$, this polynomial's degree is at most $2\alpha$, and hence there exists a coefficient vector $\vc_i \in \R^{2\alpha}$ encoding this polynomial.

Let $\mC \in \R^{n \times 2\alpha}$ be the matrix resulting from stacking such coefficient vectors for each of the $n$ nodes. Consider $\mP = \mV \mC \in \R^{n \times n}$: $\mP_{i,j}$ is $0$ if $\mB_{i,j} = 1$ and negative otherwise. Then $(\mP \circ \mP^\top)_{i,j}$ is $0$ when either $\mB_{i,j}=1$ or $(\mB^\top)_{i,j}=1$ and positive otherwise; equivalently, since $\mA = \mB + \mB^\top$, $(\mP \circ \mP^\top)_{i,j}=0$ iff $\mA_{i,j}=1$. Take any positive $\epsilon$ less than the smallest positive entry of $\mP \circ \mP^\top$. Letting $\mJ$ be an all-ones matrix, define $\mM = \epsilon \mJ - (\mP \circ \mP^\top)$. Note that $\mM_{i,j}>0$ if $\mA = 1$ and $\mM_{i,j}<0$ if $\mA = 0$, that is, $\mM = H(\mA)$ as desired. Since $\text{rank}(\mJ)=1$ and $\text{rank}(\mP) \leq 2\alpha$, by the bound on the rank of entrywise products of matrices, the rank of $\mM$ is at most $(2\alpha)^2 + 1$.
\end{proof}

\paragraph{Exact representation with community factorization} LPCA factors $\mX,\mY \in \R^{n \times k}$ can be processed into nonnegative factors $\mB \in \mathbb{R}_+^{n \times k_B}$ and $\mC \in \mathbb{R}_+^{n \times k_C}$ such that $k_B + k_C = 6k$ and
\begin{equation}\label{eqn:bbcc_xyyx}
    \mB \mB^\top - \mC \mC^\top = \tfrac{1}{2} \left( \mX\mY^\top + \mY\mX^\top \right).
\end{equation}
Observe that the left-hand side can only represent symmetric matrices, but $\mX \mY^\top$ is not necessarily symmetric even if $H(\mX \mY^\top)=\mA$ for a symmetric $\mA$. For this reason, we use a symmetrization: let $\mL = \tfrac{1}{2} \left( \mX\mY^\top + \mY\mX^\top \right)$. Note that $H(\mL) = H(\mX \mY^\top)$, so if $\mX \mY^\top$ constitutes an exact representation of $\mA$ in that $H(\mX \mY^\top) = \mA$, so too do both expressions for $\mL$ in Equation~\ref{eqn:bbcc_xyyx}.
Pseudocode for the procedure of constructing $\mB,\mC$ given $\mX,\mY$ is given in Algorithm~\ref{alg:init_constrained}. The concept of this algorithm is to first separate the logit matrix $\mL$ into a sum and difference of rank-$1$ components via eigendecomposition.
Each of these components can be written as $+\vv \vv^\top$ or $-\vv \vv^\top$ with $\vv \in \mathbb{R}^n$, where the sign depends on the sign of the eigenvalue. Each component is then separated into a sum and difference of three outer products of nonnegative vectors, via Lemma~\ref{lem:rankone_to_nonneg} below.
\begin{lemma}~\label{lem:rankone_to_nonneg}
Let $\phi : \R \rightarrow \R$ denote the ReLU function, i.e., $\phi(z) = \max\{z,0\}$. 
For any vector $\vv$,
\begin{align*}
    \vv \vv^\top = 2 \phi(\vv) \phi(\vv)^\top + 2 \phi(-\vv) \phi(-\vv)^\top - |\vv| |\vv|^\top .
\end{align*}
\end{lemma}
\begin{proof}
Take any $\vv \in \mathbb{R}^k$. Then
\begin{align*}
    \vv \vv^\top = & ~ ( \phi(\vv) - \phi(-\vv) ) \cdot ( \phi(\vv)^\top - \phi(-\vv)^\top ) \\
    = & ~ \phi(\vv) \phi(\vv)^\top + \phi(-\vv) \phi(-\vv)^\top - \phi(\vv) \phi(-\vv)^\top - \phi(-\vv) \phi(\vv)^\top \\
    = & ~ 2 \phi(\vv) \phi(\vv)^\top + 2 \phi(-\vv) \phi(-\vv)^\top - ( \phi(\vv) + \phi(-\vv) ) \cdot ( \phi(\vv) + \phi(-\vv) )^\top \\
    = & ~  2 \phi(\vv) \phi(\vv)^\top + 2 \phi(-\vv) \phi(-\vv)^\top - |\vv| |\vv|^\top,
\end{align*}
where the first step follows from $\vv= \phi(\vv)- \phi(-\vv)$, and the last step from $|\vv| = \phi(\vv) + \phi(-\vv)$.
\end{proof}
%
% Algorithm~\ref{alg:init_constrained} follows from Lemma~\ref{lem:rankone_to_nonneg} and constitutes a constructive proof of the following Lemma~\ref{thm:nneg_rank} and Theorem~\ref{thm:nneg_from_lpca}.
Algorithm~\ref{alg:init_constrained} follows from Lemma~\ref{lem:rankone_to_nonneg} and constitutes a constructive proof of the following theorem:% Theorem~\ref{thm:nneg_from_lpca}.
%
% \begin{lemma}[Nonnegative Factorization of Symmetric Rank-$k$ Matrices]\label{thm:nneg_rank}
% Given a symmetric rank-$k$ matrix $\mL \in \mathbb{R}^{n \times n}$, there exist nonnegative matrices $\mB \in \mathbb{R}_+^{n \times k_B}$ and $\mC \in \mathbb{R}_+^{n \times k_C}$ such that $k_B + k_C = 3k$ and $\mB \mB^\top - \mC \mC^\top = \mL$.
% \end{lemma}
%
\begin{theorem}[Exact Community Factorization from Exact LPCA Factorization]\label{thm:nneg_from_lpca}
Given a symmetric matrix $\mA \in \{0,1\}$ and $\mX,\mY \in \R^{n \times k}$ such that $\mA = H(\mX \mY^\top)$, there exist nonnegative matrices $\mB \in \mathbb{R}_+^{n \times k_B}$ and $\mC \in \mathbb{R}_+^{n \times k_C}$ such that $k_B + k_C = 6k$ and $\mA = H(\mB \mB^\top - \mC \mC^\top)$.
\end{theorem}

% Note that Theorem~\ref{thm:nneg_rank} and Algorithm~\ref{alg:init_constrained} show that unconstrained factors $\mX, \mY$ for the LPCA model can be processed into symmetric and nonnegative factors $\mB,\mC$ for our model without any approximation error, at the cost of increasing the factorization rank; while this exactness is of theoretical interest (discussed in Section~\ref{sec:theory}), for practical applications, some approximation error in the initialization is unimportant, and compactness of the representation is of greater interest.
% Therefore, as a heuristic for the initialization, of the $3k$ communities generated by this step, we keep the top $k$ which are most impactful on the edge logits, as ranked by the $L_2$ norms of the columns of $\mB$ and $\mC$. Now we have $\mB \in \mathbb{R}_+^{n \times k_B}$ and $\mC \in \mathbb{R}_+^{n \times k_C}$ such that $k_B + k_C = k$. This concludes the initialization; these $k$ communities can then be directly optimized with Algorithm~\ref{alg:fit_bbcc} as discussed at the start of the section.

\begin{algorithm}
    \caption{Converting LPCA Factorization to Community Factorization}
    \label{alg:init_constrained}
    \textbf{input} logit factors $\mX,\mY \in \mathbb{R}^{n \times k}$ \\
    \textbf{output} $\mB \in \R_+^{n \times k_B}$ and $\mC \in \mathbb{R}_+^{n \times k_C}$ such that $k_B+k_C=6k$ and\par\phantom{\textbf{output}} $\mB \mB^\top - \mC \mC^\top = \tfrac{1}{2} \left( \mX\mY^\top + \mY\mX^\top \right)$
    \begin{algorithmic}[1]
    \State Set $\mQ\in\mathbb{R}^{n \times 2k}$ and $\bm{\lambda} \in \mathbb{R}^{2k}$ by truncated eigendecomposition such that \par $\mQ \times \diag(\bm{\lambda}) \times \mQ^\top = \tfrac{1}{2} ( \mX\mY^\top + \mY\mX^\top )$
    \State $\mB^* \gets \mQ^+ \times \diag(\sqrt{+\bm{\lambda^+}})$, where $\bm{\lambda^+}$, $\mQ^+$ are the positive eigenvalues/vectors
    \State $\mC^* \gets \mQ^- \times \diag(\sqrt{-\bm{\lambda^-}})$, where $\bm{\lambda^-}$, $\mQ^-$ are the negative eigenvalues/vectors
    \State $\mB \gets \begin{pmatrix} \sqrt{2} \phi(\mB^*); & \sqrt{2} \phi(- \mB^* ); & | \mC^* | \end{pmatrix}$ \Comment{\textcolor{gray}{$\phi$ and $|\cdot|$ are entrywise ReLU and absolute value}}
    \State $\mC \gets \begin{pmatrix} \sqrt{2} \phi(\mC^*); & \sqrt{2} \phi(-\mC^*); & |\mB^*| \end{pmatrix}$
    \State \textbf{return} $\mB,\mC$
    \end{algorithmic}
\end{algorithm}

% \textsc{SymNMF} and \textsc{BigClam}, among other models for undirected graphs, assume network homophily, which precludes low-rank representation of networks with heterophily.
% We first show that our model is highly expressive in that it can capture arbitrary homophilous and heterophilous structure: using a result from \citet{chanpuriya2020node}, we show that our model can exactly reconstruct any graph using an embedding length that is linear in the graph's maximum degree.
%
As stated in the introduction to this section, Theorem~\ref{thm:nneg_from_lpca} extends any upper bound on the exact factorization dimensionality from the LPCA model to our community-based model. That is, up to a constant factor, the bound in terms of max degree from Theorem~\ref{thm:lpca_exact_maxdeg} and the bound in terms of arboricity from Theorem~\ref{thm:lpca_exact_maxdeg_arboricity} also apply to our model; for brevity, we state just the latter here.
\begin{corollary}[Exact Community Factorization for Bounded-Arboricity Graphs]\label{thm:exact_nonneg_arbor}
Let $\mA \in \{0,1\}^{n \times n}$ be the adjacency matrix of an undirected graph $G$ with arboricity $\alpha$. Then there exist nonnegative embeddings $\mB \in \mathbb{R}_+^{n \times k_B}$ and $\mC \in \mathbb{R}_+^{n \times k_C}$ such that $k_B + k_C = 6(4\alpha^2+1)$ and \mbox{$\mA = H(\mB \mB^\top - \mC \mC^\top)$}.
\end{corollary}
Note that Corollary~\ref{thm:exact_nonneg_arbor} is purely a statement about the capacity of our model; 
Theorem~\ref{thm:lpca_exact_maxdeg_arboricity} stems from a constructive proof based on polynomial interpolation, and therefore so too does this corollary. We do not expect this factorization to be informative about the graph's latent structure. In the following Section~\ref{sec:exp}, we will fit the model with an entirely different algorithm for downstream applications.

\section{Experiments}\label{sec:exp}

We now present a training algorithm to fit our model, then evaluate our method on a benchmark of five real-world networks. These are fairly common small to mid-size datasets ranging from around 1K to 10K nodes; for brevity, we defer the statistics and discussion of these datasets, including how some of them exhibit heterophily, to Appendix~\ref{app:datasets}.

\subsection{Training Algorithm} \label{sec:train}

Given an input graph $\mA \in \{0,1\}^{n \times n}$, we find low-rank nonnegative matrices $\mB$ and $\mC$ such that the model produces \mbox{$\tilde{\mA}=\sigma(\mB \mB^\top - \mC \mC^\top) \in (0,1)^{n \times n}$} as in Equation~\ref{eqn:bbcc} which approximately matches $\mA$.
In particular, we train the model to minimize the sum of binary cross-entropies of the link predictions over all pairs of nodes:
\begin{equation}~\label{eqn:ew_loss}
    R = -\sum\left( \mA \log(\tilde{\mA}) + (1-\mA) \log(1-\tilde{\mA}) \right),
\end{equation}
where $\sum{}$ denotes the scalar summation of all entries in the matrix. We fit the parameters by gradient descent over this loss, as well as $L_2$ regularization of the factors $\mB$ and $\mC$, subject to the nonnegativity of $\mB$ and $\mC$.
This algorithm is fairly straightforward; pseudocode is given in Algorithm~\ref{alg:fit_bbcc}.
This is quite similar to the training algorithm of \cite{chanpuriya2020node}, but in contrast to that work, which only targets an exact fit, we explore the expression of graph structure in the factors and their utility in downstream tasks. Regularization of the factors is implemented to this end to avoid overfitting.
Though we outline a non-stochastic version of the training algorithm, it generalizes straightforwardly to a stochastic version, i.e., by sampling links and non-links for the loss function.
%

% Although random initialization of $\mB$ and $\mC$ is possible, and was used to produce Figures~\ref{fig:synth_true}-\ref{fig:synth_feats}, we propose a scheme for more principled initialization these factors. 
% As one benefit, this scheme can allow for selecting a total number of clusters $k$, then automatically setting a split of homophilous/heterophilous clusters $k_B$/$k_C$ such that $k_B+k_C=k$. 
% Additionally, the scheme illuminates the connection of our model from Equation~\ref{eqn:bbcc} with existing models, yielding theoretical results discussed in Section~\ref{sec:theory}.
% As an outline, our procedure works in two steps: first, we fit the logistic principal components analysis (LPCA) model to the input graph, yielding unconstrained factors $\mX$ and $\mY$; then, we process these unconstrained factors into nonnegative initializations for $\mB$ and $\mC$.

\begin{algorithm}
    \caption{Fitting the Constrained Model}
    \label{alg:fit_bbcc}
    \textbf{input} adjacency matrix $\mA \in \{0,1\}^{n \times n}$, regularization weight $\lambda\geq 0$, number of iterations $I$, \par\phantom{\textbf{input} }number of homophilous/heterophilous communities $k_B/k_C$ \\
    % \par\hspace{25pt} initial homo- and hetero- philous assignments $\mB \in \mathbb{R}_+^{n \times k_B}$ and $\mC \in \mathbb{R}_+^{n \times k_C}$
    \textbf{output} fitted factors $\mB \in \mathbb{R}_+^{n \times k_B}$ and $\mC \in \mathbb{R}_+^{n \times k_C}$ such that $\sigma(\mB\mB^\top - \mC\mC^\top) \approx \mA$
    \begin{algorithmic}[1] % The number tells where the line numbering should start
    \State Initialize $\mB, \mC$ by setting entries to independent samples of $\text{Unif}(0,\sfrac{1}{\sqrt{k_B}}), \text{Unif}(0,\sfrac{1}{\sqrt{k_C}})$
	\For{$i \gets 1$ to $I$}                    	       
		\State $\tilde{\mA} \gets \sigma(\mB\mB^\top - \mC\mC^\top)$ %\newline \Comment{\textcolor{gray}{reconstructed adjacency matrix}}
		\State $R \gets -\sum\left( \mA \log(\tilde{\mA}) + (1-\mA) \log(1-\tilde{\mA}) \right) $ %\newline \Comment{\textcolor{gray}{cross-entropy loss}}
		\State $R \gets R + \lambda \left(\Vert \mB \Vert_F^2 + \Vert \mC \Vert_F^2\right) $ %\newline \Comment{\textcolor{gray}{regularization loss}}
		\State Calculate $\partial_{\mB,\mC} R$ via differentiation through Steps 2 to 4
		\State Update $\mB,\mC$ to minimize $R$ using $\partial_{\mB,\mC} R$, subject to $\mB,\mC \geq 0$
	\EndFor
    \State \textbf{return} $\mB,\mC$
    \end{algorithmic}
\end{algorithm}

\paragraph{Implementation details} Our implementation uses PyTorch~\citep{NEURIPS2019_9015} for automatic differentiation and minimizes loss using the SciPy~\citep{scipy} implementation of the L-BFGS~\citep{liu1989limited,zhu1997algorithm} algorithm with default hyperparameters and up to a max of 200 iterations of optimization. We set regularization weight $\lambda = 10$ as in \citet{yang2013overlapping}. 
% We include code in the form of a Jupyter notebook~\citep{PER-GRA:2007} demo. % in the supplemental material.
% The code also contains stochastic (i.e., more scalable) versions of the optimization functions.

\subsection{Results} \label{sec:res}

% \begin{table*}
% \small
% \centering
% \caption{\label{tab:datasets} Datasets used in our experiments. As in \citet{sun2019vgraph}, for {\sc YouTube} and {\sc Amazon}, we take only nodes which participate in at least one of the largest $5$ ground-truth communities. Note that degeneracy is an upper bound on arboricity.}
% % \vspace{2mm}
% \begin{tabular}{llrrrrr}
%         \toprule
%         \textbf{Name} & \textbf{Reference} & \textbf{Nodes} & \textbf{Edges} & \textbf{Labels} & \textbf{Max Degree} & \textbf{Degeneracy} \\
%         \midrule
%         {\sc Blog} & \citet{tang2009relational}   & 10,312 & 333,983 & 39 & 3,992 & 114\\
%         {\sc YouTube} & \citet{yang2015defining}  & 5,346 & 24,121 & 5 & 628 & 19\\
%         {\sc POS} & \citet{qiu2018network}  & 4,777 & 92,406 & 40 & 3,644 & 49\\
%         {\sc PPI} & \citet{breitkreutz2007biogrid}  & 3,852 & 76,546 & 50 & 593 & 29\\
%         {\sc Amazon} & \citet{yang2015defining}  & 794 & 2,109 & 5 & 29 & 6\\
%         \bottomrule
% \end{tabular}
% \end{table*}

\begin{figure*}[h!]
\centering
\includegraphics[height=1.2in]{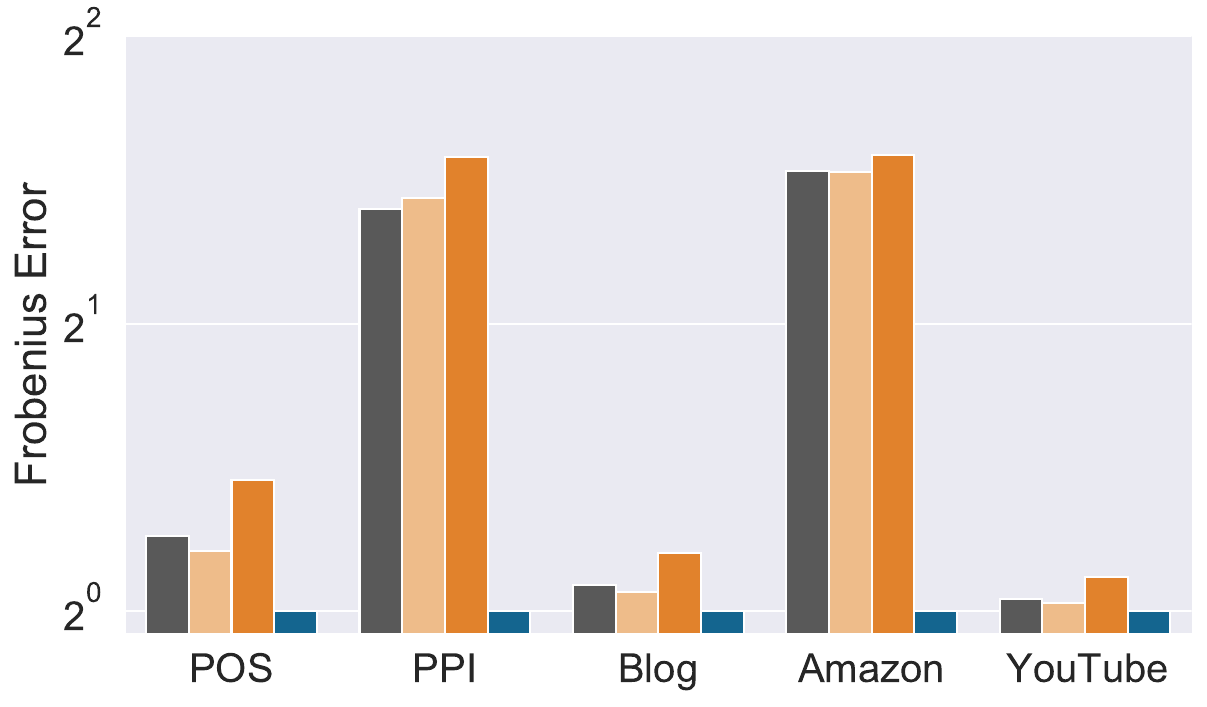}
\hspace{5pt}
\includegraphics[height=1.2in]{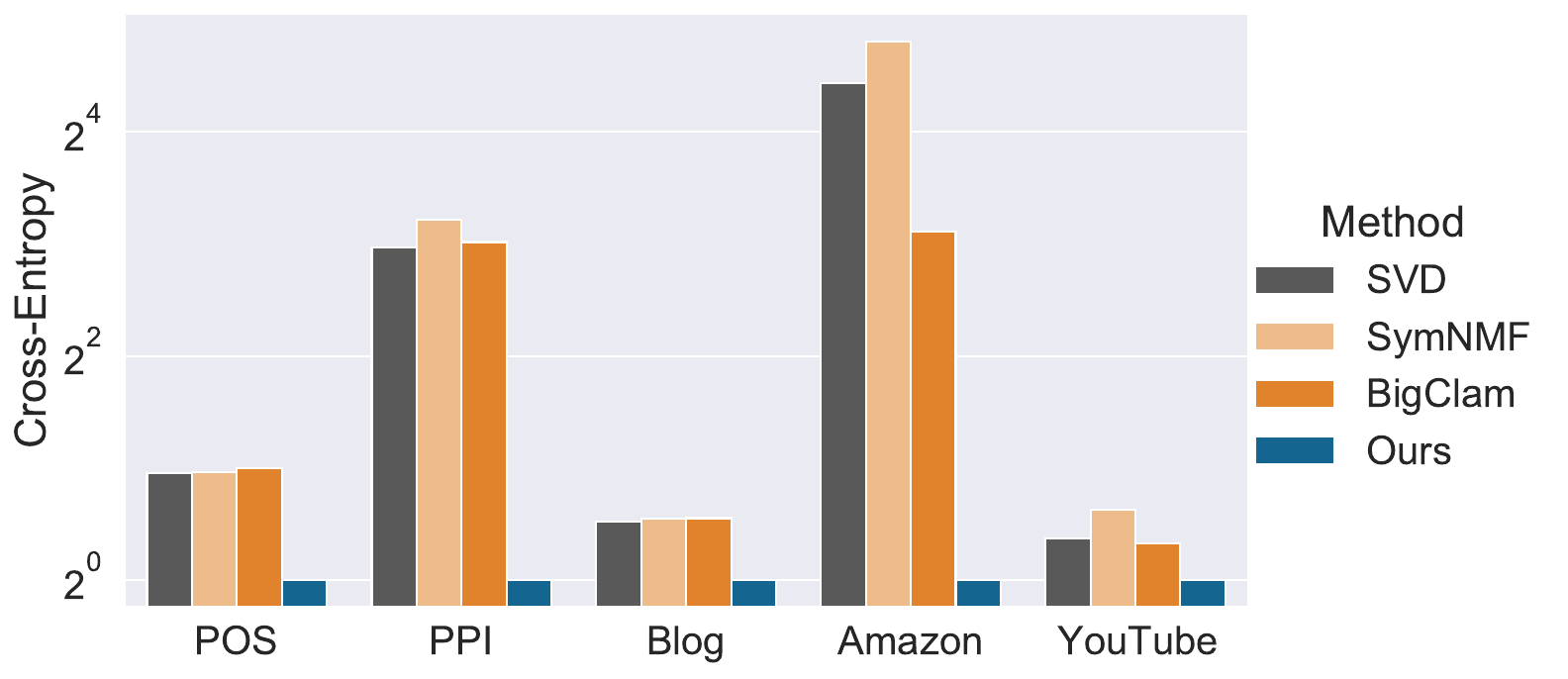}
\caption{Reconstruction error on real-world networks, relative to our model's error.}
\label{fig:real_recons}
\end{figure*}

\paragraph{Expressiveness}
First, we investigate the expressiveness of our generative model, that is, the fidelity with which it can reproduce an input network. 
In Section~\ref{sec:intro}, we used a simple synthetic network to show that our model is more expressive than others due to its ability to represent heterophilous structures in addition to homophilous ones.
We now evaluate the expressiveness of our model on real-world networks.
As with the synthetic graph, we fix the number of communities or singular vectors, fit the model, then evaluate the reconstruction error. In Figure~\ref{fig:real_recons}, we compare the results of our model with those of SVD, \textsc{BigClam} (which is discussed in detail in Section~\ref{sec:related-work}), and \textsc{SymNMF}~\citep{kuang2015symnmf}. \textsc{SymNMF} simply factors the adjacency matrix as $\mA \approx \mH \mH^\top$, where $\mH \in \mathbb{R}_{+}^{n \times k}$; note that, like SVD, \textsc{SymNMF} does not necessarily output a matrix whose entries are probabilities (i.e., bounded in $[0,1]$), and hence it is not a graph generative model like ours and \textsc{BigClam}.

% \begin{figure*}[h!]
% \centering
% \includegraphics[height=1.25in]{images/real/real_recon_frob.pdf}
% \hspace{5pt}
% \includegraphics[height=1.25in]{images/real/real_recon_ce.pdf}
% \caption{Error when reconstructing real-world graphs with several models.
% Frobenius error is normalized as in Figure~\ref{fig:synth_recons}; cross-entropy is normalized by the number of entries of the matrix ($n^2$).}
% \label{fig:real_recons}
% \end{figure*}

For each method, we fix the number of communities or singular vectors at the ground-truth number. For this experiment only, we are not concerned with learning the latent structure of the graph; the only goal is accurate representation of the network with limited parameters. So, for a fair comparison with SVD, we do not regularize the training of the other methods. Our method consistently has the lowest reconstruction error, both in terms of Frobenius error and entrywise cross-entropy (Equation~\ref{eqn:ew_loss}). Interestingly, we find the most significant improvement exactly on the three datasets which have been noted to exhibit significant heterophily: \textsc{POS}, \textsc{PPI}, and \textsc{Amazon}.

% \begin{figure}[ht]
% \begin{minipage}[b]{.45\textwidth}
% \centering
% \includegraphics[width=1\textwidth]{plots/real/real_gt.png}
% \caption{Similarity of recovered communities to ground-truth labels of real-world datasets.}
% \end{minipage}
% \hfill
% \begin{minipage}[b]{.45\textwidth}
% \centering
% \includegraphics[width=1\textwidth]{plots/real/real_linkpred.png}
% \caption{Accuracy of link prediction on real-world datasets.}
% \end{minipage}
% \end{figure}

\begin{figure}[h!]
\centering
\includegraphics[height=1.3in]{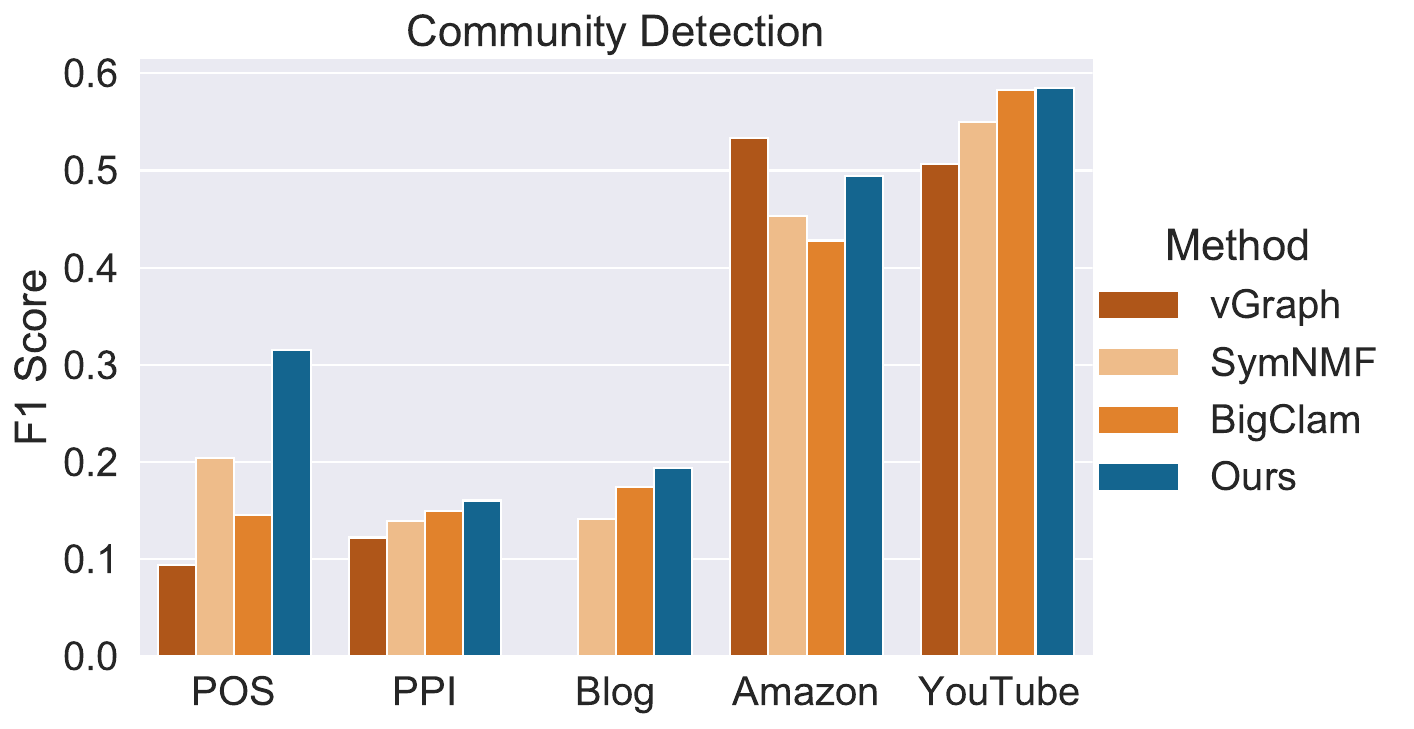}
% \hfill
\hspace{5pt}
\includegraphics[height=1.3in]{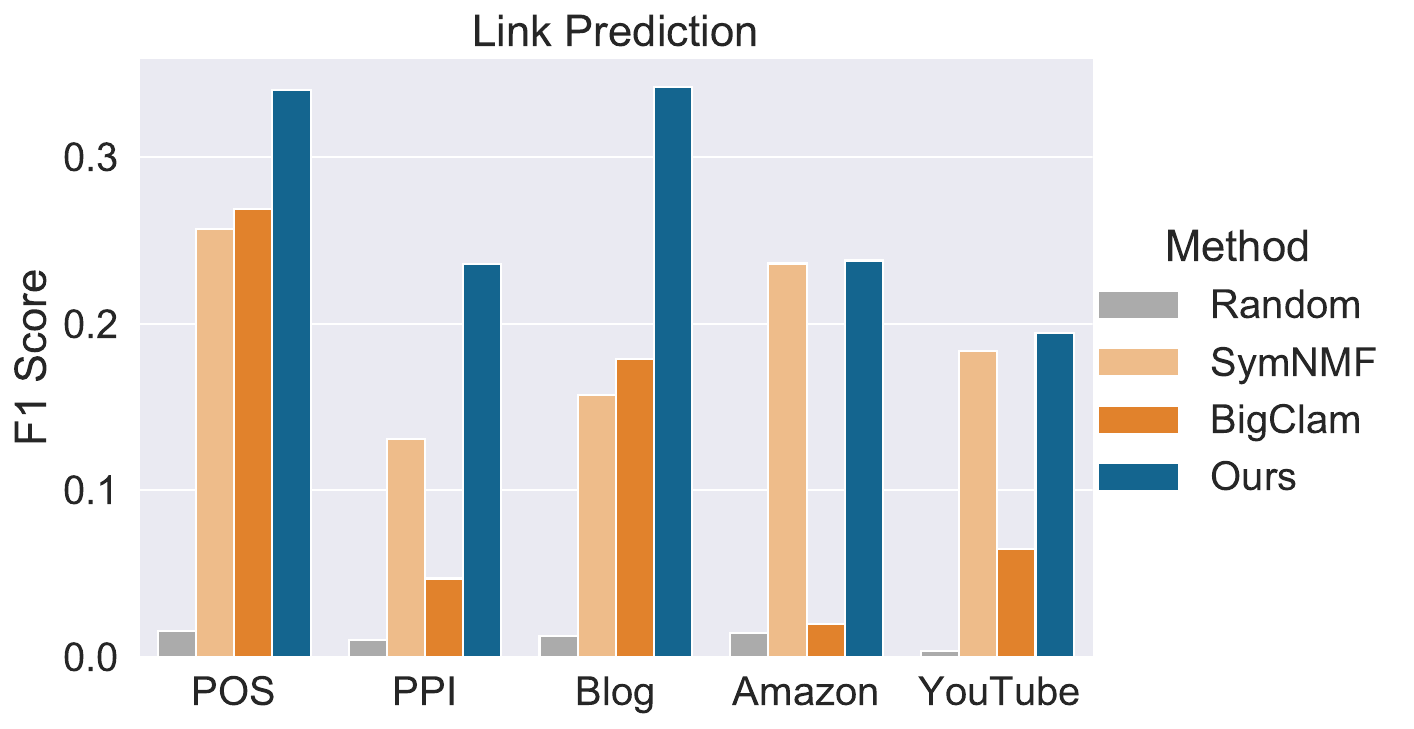}
% \vspace{-2mm}
\caption{Left: Similarity of recovered communities to ground-truth labels of real-world datasets. We are unable to run the authors' implementation of \textsc{vGraph} on \textsc{Blog} with limited memory. \mbox{Right: Accuracy} of link prediction on real-world datasets.}
\label{fig:real_gt_linkpred}
\end{figure}

% \vspace{-1pt}
\paragraph{Similarity to ground-truth communities}\label{sec:exp-ground}
To assess the interpretability of clusters generated by our method, we evaluate the similarity of these clusters to ground-truth communities (i.e., class labels), and we compare other methods for overlapping clustering.
We additionally compare to another recent but non-generative approach, the \textsc{vGraph} method of \citet{sun2019vgraph}, which is based on link clustering; the authors found their method to generally achieve state-of-the-art results in this task.
For all methods, we set the number of communities to be detected as the number of ground-truth communities.
We report F1-Score as computed in \citet{yang2013overlapping}. See Figure~\ref{fig:real_gt_linkpred} (left): the performance of our method is competitive with \textsc{SymNMF}, \textsc{BigClam}, and vGraph.

% \vspace{-1pt}
\paragraph{Interpretable link prediction}\label{sec:linkpred}
We assess the predictive power of our generative model on the link prediction task.
As discussed in Section~\ref{sec:model}, the link probabilities output by our model are interpretable in terms of a clustering of nodes that it generates; we compare results with our method to those with other models which permit similar interpretation, namely \textsc{BigCLAM} and \textsc{SymNMF}.
We randomly select 10\% of node pairs to hold out, fit the models on the remaining 90\%, then use the trained models to predict links between node pairs in the held out 10\%.
As a baseline, we also show results for randomly predicting link or no link with equal probability.
See Figure~\ref{fig:real_gt_linkpred} (right).
The performance of our method is competitive with or exceeds that of the other methods in terms of F1 Score.

% \vspace{-3mm}
% \section{Conclusion, Limitations, and Broader Impact} \label{sec:conclusion}
\section{Conclusion} \label{sec:conclusion}
We introduce a community-based graph generative model based on symmetric nonnegative matrix factorization which is capable of representing both homophily and heterophily. We expand on prior work guaranteeing exact representation of bounded max degree graphs with a new, more applicable guarantee for bounded arboricity graphs, and we show that both the prior bound and our new one apply to our more interpretable graph model.
We illustrate our model's capabilities with experiments on a synthetic motivating example. Experiments on real-world networks show its effectiveness on several key tasks.
More broadly, our results suggest that incorporating heterophily into models and methods for networks can improve both theoretical grounding and overall empirical performance, while maintaining simplicity and interpretability.
A deeper understanding of the expressiveness of both nonnegative and arbitrary low-rank logit models for graphs
% , as well as convergence properties of training algorithms,
is an interesting future direction.

\bibliographystyle{iclr2021_conference}
\bibliography{iclr2021_conference}

%%%%%%%%%%%%%%%%%%%%%%%%%%%%%%%%%%%%%%%%%%%%%%%%%%%%%%%%%%%%

\clearpage
\appendix

\section{Appendix}

% \subsection{Proof of Theorem~\ref{thm:exact_nonneg}} \label{app:proof_exact_nonneg}

% \begin{proof}
% Theorem~\ref{thm:lpca_exact_maxdeg} guarantees the existence of matrices $\mX,\mY \in \R^{n \times (2c+1)}$ such that $(\mX \mY^\top)_{ij} > 0$ if $\mA_{ij} = 1$ and $(\mX \mY^\top)_{ij} < 0$ if $\mA_{ij} = 0$. Let $\mL = \tfrac{1}{2}(\mX \mY^\top + \mY \mX^\top)$, the symmetrization of $\mX \mY^\top$. Since $\mA$ is symmetric, it still holds that $\mL_{ij} > 0$ if $\mA_{ij} = 1$ and $\mL_{ij} < 0$ if $\mA_{ij} = 0$; further, as the sum of two rank-$(2c+1)$ matrices, the rank of $\mL$ is at most $2\cdot (2c+1)$. Finally, by Theorem~\ref{thm:nneg_rank} and Equation~\ref{eqn:bbcc_to_vwv}, with $k=3\cdot 2\cdot (2c+1) = 12c+6$, there exist $\mV \in [0,1]^{n\times k}$ and diagonal $\mW \in \mathbb{R}^{k \times k}$ such that $\mV \mW \mV^\top = \mL$, meaning still $(\mV \mW \mV^\top)_{ij} > 0$ if $\mA_{ij} = 1$ and $(\mV \mW \mV^\top)_{ij} < 0$ if $\mA_{ij} = 0$.
% Since $\lim_{z\to -\infty} \sigma(z) = 0$ and $\lim_{z\to +\infty} \sigma(z) = 1$, it follows that
% \[ \lim_{s\to \infty} \sigma \left(\mV (s\mW) \mV^\top \right) = \lim_{s\to \infty} \sigma \left(s \mV \mW \mV^\top \right) = \mA, \]
% that is, $\mW$ can be scaled larger to match $\mA$ arbitrarily closely.
% \end{proof}

\subsection{COT Graph Exact Representation} \label{app:cot}

As a theoretical demonstration of the capability of our model to learn latent structure, we additionally show that our model can exactly represent a natural family of graphs, which exhibits both homophily and heterophily, with small $k$ and interpretably. The family of graphs is specified below in Definition~\ref{def:cot}; roughly speaking, nodes in such graphs share an edge iff they coparticipate in some number of homophilous communities and don't coparticipate in a number of heterophilous communities. For example, the motivating graph described in Section~\ref{sec:intro} would be an instance of such a graph if an edge occurs between two users iff the two users are from the same city and have different genders.
\begin{definition}[Community Overlap Threshold (COT) Graph] \label{def:cot}
An unweighted, undirected graph
%  on $n$ nodes with adjacency matrix $\mA \in \{0,1\}^{n \times n}$
whose edges are determined by an overlapping clustering and a ``thresholding'' integer $t \in \mathbb{Z}$ as follows: for each vertex $i$, there are two latent binary vectors $\vb_i \in \{0,1\}^{k_b}$ and $\vc_i \in \{0,1\}^{k_c}$, and there is an edge between vertices $i$ and $j$ iff \mbox{$\vb_i \cdot \vb_j - \vc_i \cdot \vc_j \geq t$}.
\end{definition}
\begin{theorem}[Compact Representation of COT Graphs] \label{thm:cot_rep}
Suppose $\mA$ is the adjacency matrix of a COT graph on $n$ nodes with latent vectors $\vb_i \in \{0,1\}^{k_b}$ and $\vc_i \in \{0,1\}^{k_c}$ for $i\in\{1,2,\dots,n\}$. Let $k = k_b + k_c$.
Then, for any $\epsilon > 0$, there exist $\mV \in [0,1]^{n\times (k+1)}$ and diagonal $\mW \in \mathbb{R}^{(k+1) \times (k+1)}$ such that $\norm{\sigma (\mV \mW \mV^\top) - \mA}_\text{F} < \epsilon$.
\end{theorem}
%
% We defer the proof to Appendix~\ref{app:proof_cot_rep}; the concept is that the latent vectors can be processed into community factors $\mB,\mC$, and the thresholding integer can be handled with an extra community.
%
\begin{proof}
Let $t$ be the thresholding integer of the graph, and let the rows of $\mB \in \{0,1\}^{n \times k_b}$ and $\mC \in \{0,1\}^{n \times k_c}$ contain the vectors $\vb$ and $\vc$ of all nodes.
% Then $\mA_{ij} = 1$ iff $(\mB \mB^\top - \mC \mC^\top)_{ij} \geq t$.
Via Equation~\ref{eqn:bbcc_to_vwv}, we can find $\mV^* \in [0,1]^{n\times k}$ and diagonal $\mW^* \in \mathbb{R}^{k \times k}$ such that $\mV^* \mW^* \mV^{*\top} = \mB \mB^\top - \mC \mC^\top$. Now let
\begin{align*}
    \mV = \begin{pmatrix} \mV^* & \bm{1} \end{pmatrix} \qquad
    \mW = \begin{pmatrix} \mW^* & 0 \\ 0 & \tfrac{1}{2}-t \end{pmatrix} .
\end{align*}
Then $(\mV \mW \mV^{\top})_{ij} = \vb_i \cdot \vb_j - \vc_i \cdot \vc_j + \tfrac{1}{2} - t$. Hence $(\mV \mW \mV^{\top})_{ij} > 0$ iff $\vb_i \cdot \vb_j - \vc_i \cdot \vc_j > t - \tfrac{1}{2}$, which is true iff $\mA_{ij} = 1$ by the assumption on the graph. Similarly, $(\mV \mW \mV^{\top})_{ij} < 0$ iff $\mA_{ij} = 0$. It follows that
\begin{equation*}
    \lim_{s\to \infty} \sigma \left(\mV (s\mW) \mV^\top \right) = \lim_{s\to \infty} \sigma \left(s \mV \mW \mV^\top \right) = \mA.\qedhere
\end{equation*}
\end{proof}

\clearpage
\subsection{Dataset Descriptions} \label{app:datasets}

As stated in Section~\ref{sec:exp}, we now briefly describe the five real-world datasets. Statistics for these datasets are given in Table~\ref{tab:datasets}.

\begin{table}[h!]
\small
\centering
\caption{Statistics of datasets used in our experiments. As in \citet{sun2019vgraph}, for {\sc YouTube} and {\sc Amazon}, we take only nodes which participate in at least one of the largest $5$ ground-truth communities. Note that degeneracy is an upper bound on arboricity.}\label{tab:datasets}
% \vspace{2mm}
\begin{tabular}{llrrrrr}
        \toprule
        \textbf{Name} & \textbf{Reference} & \textbf{Nodes} & \textbf{Edges} & \textbf{Labels} & \textbf{Max Degree} & \textbf{Degeneracy} \\
        \midrule
        {\sc Blog} & \citet{tang2009relational}   & 10,312 & 333,983 & 39 & 3992 & 114\\
        {\sc YouTube} & \citet{yang2015defining}  & 5,346 & 24,121 & 5 & 628 & 19\\
        {\sc POS} & \citet{qiu2018network}  & 4,777 & 92,406 & 40 & 3644 & 49\\
        {\sc PPI} & \citet{breitkreutz2007biogrid}  & 3,852 & 76,546 & 50 & 593 & 29\\
        {\sc Amazon} & \citet{yang2015defining}  & 794 & 2,109 & 5 & 29 & 6\\
        \bottomrule
\end{tabular}
\end{table}

\textbf{\textsc{Blog}} is a social network of relationships between online bloggers; the node labels represent interests of the bloggers.
Similarly, \textbf{\textsc{YouTube}} is a social network of YouTube users, and the labels represent groups that the users joined.

\textbf{\textsc{POS}} is a word co-occurrence network: nodes represent words, and there are edges between words which are frequently adjacent in a section of the Wikipedia corpus. Each node label represents the part-of-speech of the word. \textbf{\textsc{PPI}} is a subgraph of the protein-protein interaction network for Homo Sapiens. Labels represent biological states. Finally, \textbf{\textsc{Amazon}} is a co-purchasing network: nodes represent products, and there are edges between products which are frequently purchased together. Labels represent categories of products.

While social networks like the former two in this list are generally dominated by homophily~\citep{mcpherson2001birds}, the latter three should exhibit significant heterophily. For co-purchasing networks like \textsc{Amazon}, depending on the product, two of the same kind of product are generally not co-purchased, e.g., Pepsi and Coke, as discussed in \citet{peysakhovich2021attract}. Though less intuitively accessible, there is also prior discussion of disassortativity in word adjacencies~\citep{foster2010edge,zweig2016word}, as well as in PPI networks~\citep{newman2002assortative,hase2010difference}.

% \begin{table}[h!]
% \small
% \centering
% % \caption{}
% % \vspace{2mm}
% \begin{tabular}{llrrr}
%         \toprule
%         \textbf{Name} & \textbf{Reference} & \textbf{Nodes} & \textbf{Edges} & \textbf{Labels} \\
%         \midrule
%         {\sc Blog} & \citet{tang2009relational}   & 10,312 & 333,983 & 39 \\
%         {\sc YouTube} & \citet{yang2015defining}  & 5,346 & 24,121 & 5 \\
%         {\sc POS} & \citet{qiu2018network}  & 4,777 & 92,406 & 40 \\
%         {\sc PPI} & \citet{breitkreutz2007biogrid}  & 3,852 & 76,546 & 50 \\
%         {\sc Amazon} & \citet{yang2015defining}  & 794 & 2,109 & 5 \\
%         \bottomrule
% \end{tabular}
% \end{table}

\end{document}

%% file: math_commands.tex
\usepackage{amsmath,amsfonts,bm}

\def\eqref#1{equation~\ref{#1}}
\def\ceil#1{\lceil #1 \rceil}

\def\1{\bm{1}}

\def\vb{{\bm{b}}}
\def\vc{{\bm{c}}}

\def\vm{{\bm{m}}}

\def\vv{{\bm{v}}}

\def\mA{{\bm{A}}}
\def\mB{{\bm{B}}}
\def\mC{{\bm{C}}}
\def\mD{{\bm{D}}}

\def\mH{{\bm{H}}}

\def\mJ{{\bm{J}}}

\def\mL{{\bm{L}}}
\def\mM{{\bm{M}}}

\def\mP{{\bm{P}}}
\def\mQ{{\bm{Q}}}

\def\mV{{\bm{V}}}
\def\mW{{\bm{W}}}
\def\mX{{\bm{X}}}
\def\mY{{\bm{Y}}}

\DeclareMathAlphabet{\mathsfit}{\encodingdefault}{\sfdefault}{m}{sl}
\SetMathAlphabet{\mathsfit}{bold}{\encodingdefault}{\sfdefault}{bx}{n}

\newcommand{\R}{\mathbb{R}}

%% file: preamble.tex
\usepackage{microtype}
\usepackage{balance}
\usepackage{booktabs}
\usepackage{tabularx}
\usepackage{wrapfig}

\usepackage{amsmath,amssymb,amsthm}

\newcommand{\eat}[1]{\ignorespaces}
\usepackage{comment}

\newcommand{\journal}[1]{} % text that is too long for short conf paper, but would be good for journal

\usepackage{tikz}
\usepackage{verbatim}
\usetikzlibrary{arrows}
\usetikzlibrary{shapes,snakes}
\usetikzlibrary{decorations.pathmorphing} % noisy shapes
\usetikzlibrary{fit}					% fitting shapes to coordinates
\usetikzlibrary{backgrounds}	% drawing the background after the foreground

\usepackage{ragged2e}
\usepackage{multirow}
\usepackage{microtype}
\usepackage{balance}
\usepackage{setspace}

\graphicspath{{./}{./graphics/}}
\newcolumntype{H}{>{\setbox0=\hbox\bgroup}c<{\egroup}@{}}

\newcolumntype{R}[1]{>{\RaggedLeft\arraybackslash}} %p{#1}}
\newcolumntype{L}[1]{>{\RaggedRight\arraybackslash}} %p{#1}}

\newtheorem{theorem}{Theorem}[section]

\newtheorem{corollary}[theorem]{Corollary}

\newtheorem{lemma}[theorem]{Lemma}
\newtheorem{definition}{Definition}

\renewcommand{\vec}[1]{\boldsymbol{\mathrm{#1}}}
\DeclareMathOperator{\diag}{diag}%
\DeclareMathOperator{\hugeE}{\mbox{\huge\raise-0.3ex\hbox{E}}}
\DeclareMathOperator{\p}{\mathbb{P}}
\DeclareMathOperator{\hugep}{\mbox{\huge\raise-0.3ex\hbox{$\p$}}}
\providecommand{\mA}{\ensuremath{\mat{A}}}
\providecommand{\mB}{\ensuremath{\mat{B}}}
\providecommand{\mC}{\ensuremath{\mat{C}}}
\providecommand{\mD}{\ensuremath{\mat{D}}}

\providecommand{\mH}{\ensuremath{\mat{H}}}

\providecommand{\mJ}{\ensuremath{\mat{J}}}

\providecommand{\mL}{\ensuremath{\mat{L}}}
\providecommand{\mM}{\ensuremath{\mat{M}}}

\providecommand{\mP}{\ensuremath{\mat{P}}}
\providecommand{\mQ}{\ensuremath{\mat{Q}}}

\providecommand{\mV}{\ensuremath{\mat{V}}}
\providecommand{\mW}{\ensuremath{\mat{W}}}
\providecommand{\mX}{\ensuremath{\mat{X}}}
\providecommand{\mY}{\ensuremath{\mat{Y}}}

\providecommand{\vb}{\ensuremath{\vec{b}}}
\providecommand{\vc}{\ensuremath{\vec{c}}}

\providecommand{\vm}{\ensuremath{\vec{l}}}

\providecommand{\vv}{\ensuremath{\vec{v}}}
\renewcommand{\vv}{\ensuremath{\vec{v}}}

\usepackage{subfigure}
\usepackage{graphbox}
\usepackage{svg}
\usepackage{nicefrac}

\DeclareMathAlphabet{\mathbcal}{OMS}{cmsy}{b}{n}
\usepackage{amsmath}
\usepackage{mathrsfs}
\usepackage{comment}
\usepackage{bm}
\usepackage{bbm}
\usepackage{amssymb}
\usepackage{enumitem}